\documentclass[conference]{IEEEtran}
\IEEEoverridecommandlockouts

\usepackage{cite}
\usepackage{amsmath,amssymb,amsfonts}
\usepackage{graphicx}
\usepackage{textcomp}
\usepackage{xcolor}
\def\BibTeX{{\rm B\kern-.05em{\sc i\kern-.025em b}\kern-.08em
    T\kern-.1667em\lower.7ex\hbox{E}\kern-.125emX}}

\usepackage{graphicx}
\usepackage{graphics}
\usepackage{subfigure} 
\usepackage{multirow}
\usepackage{float}
\usepackage{booktabs}
\usepackage{amsthm}
\theoremstyle{definition}

\newtheorem{Theo}{Theorem}

\newtheorem{Def}{Definition}
\newenvironment{pro}{\emph{Proof:}}{\hfill $\square$\par}

\newcommand{\etal}{\textit{et al}.}
\newcommand{\ie}{\textit{i}.\textit{e}.}
\newcommand{\eg}{\textit{e}.\textit{g}.}
\newcommand{\etc}{\textit{etc.}}

\usepackage{amssymb}
\usepackage{calligra}
\usepackage{algorithm}
\usepackage{algorithmicx}
\usepackage{algpseudocode} 
\usepackage{amsmath}

\begin{document}

\title{RDP: Ranked Differential Privacy for Facial Feature Protection in Multiscale Sparsified Subspace
}

\author{\IEEEauthorblockN{Lu Ou}
\IEEEauthorblockA{\textit{Hunan University}\\
Changsha, China \\
oulu9676@gmail.com}
\and
\IEEEauthorblockN{Shaolin Liao}
\IEEEauthorblockA{\textit{IIT}\\
Chicago, USA \\
sliao5@iit.edu}
\and
\IEEEauthorblockN{Shihui Gao}
\IEEEauthorblockA{\textit{Hunan University}\\
Changsha, China \\
gaoqaq@hnu.edu.cn}
\and
\IEEEauthorblockN{Guandong Huang}
\IEEEauthorblockA{\textit{State Grid Hunan Materials Company}\\
Changsha, China \\
huanggd@hnu.edu.cn}
\and
\IEEEauthorblockN{Zheng Qin}
\IEEEauthorblockA{\textit{Hunan University}\\
Changsha, Country \\
zqin@hnu.edu.cn}
}

\maketitle

\begin{abstract}
With the widespread sharing of personal face images in applications' public databases, face recognition systems faces real threat of being breached by potential adversaries who are able to access users' face images and use them to intrude the face recognition systems. Although large amounts of work on facial features privacy preserving methods have been proposed, it is difficult to provide accurate and lightweight protection without degrading the visualization quality of face images, or data utility, under the constraint of given privacy budget for effective facial features protection. In this paper, we propose a novel privacy protection method in the multiscale sparsified feature subspaces to protect sensitive facial features, by taking care of the influence or weight ranked feature coefficients on the privacy budget, named ``Ranked Differential Privacy (RDP)''. After the multiscale feature decomposition, the lightweight Laplacian noise is added to the dimension-reduced sparsified feature coefficients according to the geometric superposition method. Then, we  rigorously prove that the RDP satisfies ${\varepsilon}_0$-Differential Privacy. After that, the nonlinear Lagrange Multiplier (LM) method is formulated for the constraint optimization problem of maximizing the utility of the visualization quality protected face images with sanitizing noise, under a given facial features privacy budget ${\varepsilon}_0$. Then, two methods are proposed to solve the nonlinear LM problem and obtain the optimal noise scale parameters: 1) the analytical Normalization Approximation (NA) method where all Laplacian noise scale parameters are normalized to the average noise scale parameter, which is good for real-time online applications; and 2) the LM optimization Gradient Descent (LMGD) numerical method to obtain the nonlinear solution through iterative updating, which is better for more accurate offline applications. Experimental results on two real-world datasets show that our proposed RDP outperforms other state-of-the-art methods: at a privacy budget of ${\varepsilon}_0 = 0.2$, the PSNR (Peak Signal-to-Noise Ratio) of the RDP is about $\sim 10$ dB higher than (10 times as high as) the highest PSNR of all compared methods.
\end{abstract}


\section{Introduction}
\begin{figure*}[htbp]
  \centering
   \includegraphics[width=0.8\linewidth]{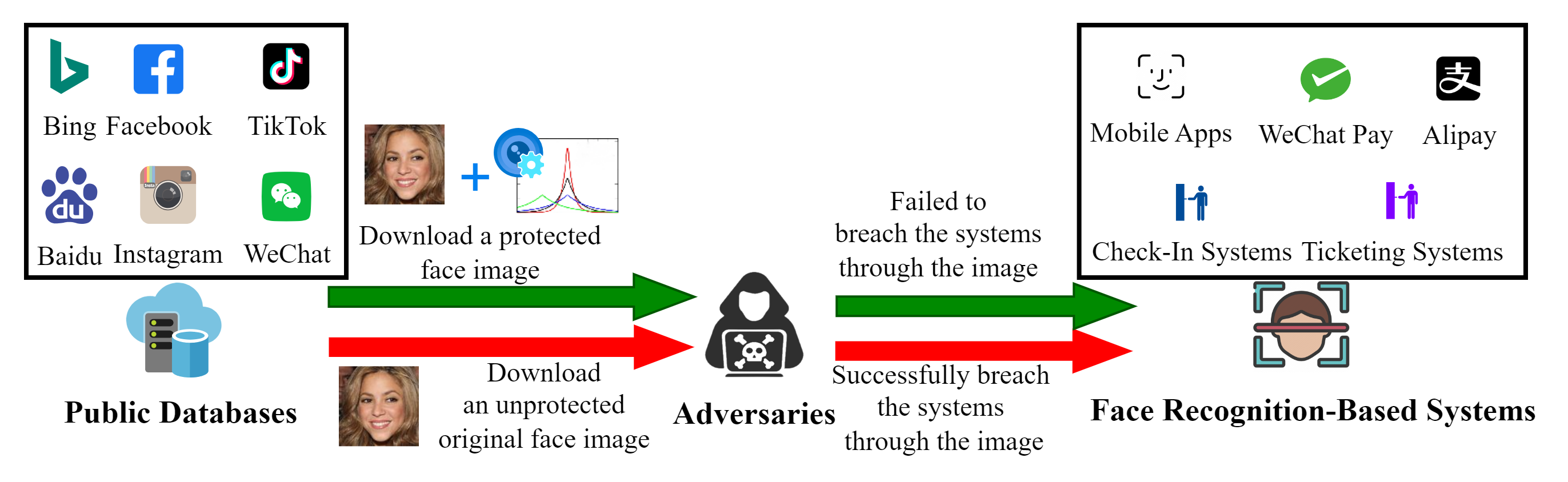}
  \caption{The process of breaching face recognition systems: for each individual's face recognition, there are the corresponding standard facial features stored in face recognition-based systems, such as face recognition check-in systems. When adversaries download the particular individual's face images from public databases, corresponding to applications, \eg, Facebook, Weibo, TikTok, or Bing, they may use the downloaded images to breach face recognition-based systems which are used by such individual. On the contrary, if all the face images are protected with a privacy protection method, adversaries will fail to achieve the threat aforementioned.}
  \label{fig:introduction}
\end{figure*}

In the current digital age, it has become a common phenomenon that individuals' static face images are shared in the public domain, such as Facebook, TikTok, Bing, \etc \cite{10.1145/3485447.3512256,9484550}. Moreover, there are universal and unique facial features with rich biometric information in such images. Existing research shows that facial features provide a solid foundation for face recognition techniques, which plays a vital role in applications such as identity authentication and access control. For example, the face recognition payment function launched by Alipay allows a user to verify his/her identity through scanning his/her face \cite{hewei2018alipay}; and it is a common identity authentication application to filter permissions using face recognition techniques \cite{10179474,9484550}. Unfortunately, adversaries could breach face recognition systems by public face images \cite{10218351,10521756,DASTMALCHI2022116755,WOS:001132069200004}. Then they can use smart devices without authorization \cite{9488737}, access private systems \cite{10179474}, and even obtain personal sensitive information \cite{10.1007/978-3-031-02444-3_11,9484550}, \etc. As shown in Figure \ref{fig:introduction}, due to the sensitive facial features in face images, once adversaries have obtained original face images in public databases, they could breach face recognition-based systems. From that, it means that facial features in public face images not only threaten personal privacy, but also may lead to a series of security threats, including identity theft, fraud, and even political and economic espionage, which has been brought up by \cite{10078250, 10203744, 10.1145/3448016.3457234}. What's more, the privacy leakage even could be caused by only one face image \cite{10179474}. All of that said, it is necessary to develop an effective facial feature protection method for public face images.

In existing research, cryptography\  \cite{10.1145/3472634.3472661,DBLP:journals/mta/WangL24}, deep learning, \  \cite{10203744,WOS:000657193000002} and Differential Privacy (DP) \cite{2020101951,10078250} are adopted to protect facial features effectively. Although cryptography-based facial feature privacy protection methods can provide strong security, they usually focus on defending against untrusted models or servers, and have expensive computation and communication overhead. Besides, deep learning\cite{WOS:000657193000002} needs to be trained on large-scale data sets to avoid privacy risks, which is more used for preserving face recognition accuracy while protecting facial features. Different from the previous two kinds of methods, DP \cite{2013The,2020101951,10078250} is a solid mathematical method that can adds sanitizing noise for protecting facial features while preserving not only the recognition accuracy but also the visual quality of images with released demand. What's more important is that DP can be optimized to maximize the data utility of face images after the sanitizing noise is added, under the constraint of a given privacy budget. Existing methods based on DP \cite{WOS:000897093900028,meng2022improving,2020101951} may be more likely to add perturbation during the training process, which comes with a large training cost.


In this paper, the data utility is the visualization quality of the noisy face images after the sanitizing noise is added. And the sensitive information is the facial features in face images that could be used to breach some face recognition-based systems. For hiding sensitive information of each individual's face image and conserving the face image's data utility as far as possible, the perturbation on the facial features should be minimized while maximizing each noisy face image's visualization quality with a given privacy budget of this image's facial features. 
And then, we propose such an optimized facial features protection method based on Ranked Differential Privacy (RDP): first, we perform the dimension-reduced sparsification of each original face image for extracting the original facial details into the multiscale feature space and obtaning sparse feature coefficients; then, the influence or weight of each sparse feature coefficient on the total privacy budget $\varepsilon_0$ is evaluated; finally, sanitizing noise of the optimal scale parameters is added to the ranked significant sparse feature coefficients by taking care of the influence or weight of each coefficient. The main contributions of this paper are as follows,
\begin{itemize}
\item First, after reducing the dimension of every face image and obtaining the sparse feature coefficients, the weights of each coefficient with respect to the facial feature vector and thus the total privacy budget $\varepsilon_0$ are evaluated and coefficients are ranked in accordance with such weights. Then, Lagrange Multiplier method (LM) constraint optimization problem is formulated: optimizing the Laplace scale parameters of all feature coefficients with significant influence or weights to  maximize the visualization quality of the privacy-preserving noisy face image, under the constraint of the total privacy budget $\varepsilon_0$ of the face image's facial features. 
\item What's more important, two methods are proposed to solve the nonlinear LM optimization problem: the Normalized Approximation (NA) analytical method by approximating all weighted Laplacian noise scale parameters to be identical, which is suitable for real-time online applications; and the iteratively updating LM optimization Gradient Descent (LMGD) numerical method, which is first proposed here to achieve more accurate offline applications.
\item Finally, experiments on two real-world datasets show that the facial features have been effectively protected, while the face image's visualization quality is only affected minimally, compared to other state-of-the-art DP methods.
\end{itemize}

\section{Related Work}

Due to facial features that existed in face images, the images could be used to breach face recognition-based systems \cite{2021Backdoors,9484550}. There is the related research on facial features protections. Then, we will review the research in recent years. 

A lot of research focus on the problem of how to reserve the efficiency of face recognition while preventing facial features in face images from being leaking for individuals' sensitivity information or face images reconstruction: Chamikara \etal \  \cite{2020101951} proposed a face recognition privacy protection scheme based on Local Differential Privacy (LDP), which calculates by perturbing the eigenface and storing the perturbed data in a third-party server;  
Ding \etal \cite{9866824} proposed a model with dual channels and multiple monitors to prevent DeepFake anti-forensic attacks for protecting facial features; Meng \etal \  \cite{meng2022improving} proposed a face privacy framework, which extracts purified clusters locally through Differentially Private Local Clustering (DPLC) method and then encourages global consensus among clients to generate more discriminative features; Wang \etal \cite{10203744} considered that facial features used by face recognition system for authentication may be attempted to reconstruct the original face image and proposed a privacy protection method based on disturbing the mapping from facial features to the image; Lin \etal \cite{10499238} proposed a privacy-preserving facial features based on a new facial recognition paradigm with a feature compensation mechanism for the challenge between the public demand for powerful face recognition and the individuals' privacy concerns; Mi \etal \cite{Mi_2024_CVPR} proposed the MinusFace facial features privacy protection framework for enhancing facial feature protection by random channel shuffling and achieving high accuracy in face recognition system. Others are more interested in preventing adversaries from attacking face recognition models or servers storing the relevant data to obtain facial features: Wang and Leng \cite{DBLP:journals/mta/WangL24} proposed an image encryption algorithm that combines face recognition and bitonal sequence technologies, and designed a row-column scrambling algorithm to scramble facial features through the characteristics of bitonal sequences; Zhuo \etal \cite{8668517} proposed an AdaBoost-based framework POR to protect the privacy of users' facial features and the service provider's key learning parameters in face recognition integrated deep learning algorithms. Additionally, there are some studies worrying about the facial features leaking of face images released to the public domain: Ji \etal \cite{WOS:000897093900028} proposed a frequency-domain DP protection method DCT-DP for facial features, with an optimized privacy budget allocation method based on the loss of the face recognition network; Croft \etal  \cite{2021Obfuscation} proposed a face obfuscation DP method for generating machine learning models, adding noise to the image pixel intensities for protecting facial features.

All above, such research effectively protects individuals' facial features in face images. However, most studies are more concerned with the trade-off between face recognition accuracy and faial features protection, or consider the unreliability of face recognition models and servers storing related data. In this paper we intend to solve how to release face images with high-quality visual effects on the premise of protecting facial features. Different from \cite{WOS:000897093900028}, we attempt to propose a lightweight perturbation method with adding less noises than \cite{2021Obfuscation} in the face images not in training model.


\section{Preliminary Work}
In this section, we will formally define the notations involved in this paper and systematically introduce models and problem statement. 

At beginning, notations are shown in Table \ref{tab:notation}.

\begin{table}[hbtp]
\caption{Notations and Definitions}
\centering
\label{tab:notation}
\begin{tabular}{cp{6cm}}
\toprule
\textbf{Symbol} & \textbf{Description}  \\
\midrule
${\bf{P}}$& A pixel matrix of a face image.\\
${\bf{C}}$& A matrix of sparsed feature coefficients for the face image.\\
${\bar P}$ & A one-dimensional pixel vector of the face image. \\ 
 $\bar C$& A one-dimensional sparsed feature coefficient vector of the face image.\\
${\bar C^ * }$& A one-dimensional sparsed feature coefficient vector after sorting the elements of $\bar C$.\\
${M_P}$ & The number of elements of $\bar C$ or ${\bar P}$.\\
${\xi _k}$ & The noise of the $k$-th element ${\bar  C_k}{^ *{} ^\prime }$ of the noisy sparsed feature coefficients vector be ${\bar C^ * }{}^\prime $,  $1 \le k \le {M_P}$.\\
$b_k$ & The Laplacian scale parameter of ${\xi _k}$, $1 \le k \le {M_P}$.\\
${{\bar F}}$ & A feature vector of the face image. \\
${M_F}$ & The number of elements of ${\bar F}$. \\
 ${\Delta}_i$ & The sensitivity of the $i$-th element ${\bar F_i}^\prime $ of the noisy feature vector $\bar F'$, $1 \le i \le {M_F}$.\\
${{w_{ik}}}$  & The weights between the $i$-th element ${\bar F_i}^\prime $ of the noisy feature vector ${{\bar F'}}$ and the $k$-th element ${\bar  C_k}{^ *{} ^\prime }$ of the ranked noisy sparsed feature coefficient vector ${\bar C^ * }{}^\prime $, $1 \le i \le {M_F}$, $1 \le k \le {M_P}$.\\ 
 $p$ &  The probability parameter of the geometric distribution, $0 < p < 1$. \\
$K$ &  The geometric number, $K \sim Geo\left( p \right)$ \\
$\varepsilon_0$ & A constant, which is given as the total privacy budget of $\bar F'$. \\
\bottomrule
\end{tabular}
\end{table}

\subsection{Models and Problem Statement}
Next, we will give the system and adversary models, and state the problem. 

\subsubsection{System Model} 
With a certain individual $\mathbb{A}$, there exist a large amount of $\mathbb{A}$'s static face images represented as a dataset ${\Omega }$ in public, which can be unrestricted observed and obtained. ${\bar F^{\left( \bar A \right)}}$ is the standard feature vector retrieved from $\mathbb{A}$'s standard face ${\bar A}$, used for authentication in some systems with face recognition techniques. For any face image $A$ visually similar to the individual $\mathbb{A}$, its feature vector is ${{\bar F}}$ in the same face recognition with ${\bar F^{\left( \bar A \right)}}$. The system identifies that individual $\mathbb{A}$ is passing and can pass through $A$, if ${{\bar F}}$ and ${\bar F^{\left( \bar A \right)}}$ meet the following matching criterion
\begin{align}
  \left| {\bar{F_i} - {\bar {F_i}^{\left( \bar A \right)}}} \right| \le {\bar R_i};\;\;i = 1,2, \cdots ,{M_F},
    \label{eq:3.1}
\end{align}
where $\bar{F_i}$ and ${\bar {F_i}^{\left( \bar A \right)}}$  are the $i$-th element of ${{\bar F}}$ and ${\bar F^{\left( \bar A \right)}}$ with ${M_F}$ elements respectively, and ${{\bar R}_i}$ is the $i$-th element of the feature identification radius vector $\bar R$ for this authentication system.


\subsubsection{Adversary and Problem}

In this paper, adversaries are considered to be interest to and have the ability to download the face image dataset ${\Omega }$ of the certain individual $\mathbb{A}$ from the public domain, and they could obtain information about authentication devices equipped with the face recognition system, which are commonly used by individual $\mathbb{A}$. Adversaries may attempt to breach these face recognition system by using such face image in ${\Omega }$, and pass the authentication to obtain $\mathbb{A}$'s privacy information.

For the face image dataset ${\Omega }$ of $\mathbb{A}$, our goal is to develop an accurate and lightweight facial feature vector privacy-preserving method $f'$ that sanitizes these images by adding noise, \ie $A' = f'\left( A \right)$ for any face image $A  \in \Omega $. Such method can ensure that the noisy face image $A'$ can prevent adversaries from breaching face recognition-based systems while having extremely high visual quality, meaning that we aim to optimize the utility of the sanitized images by maximizing images' visualization quality while providing privacy protection for facial features in face images.

\subsection{Basic Concepts}
In this section, we give some preliminary definitions for DP. 


First, in the terminology of DP, we define a pair of neighboring face images, expressed in \textbf{Definition \ref{def:neighborhood_face_image}}.

\begin{Def}[Neighboring Face Images]\label{def:neighborhood_face_image}
If a face image $A$ belongs to the individual $\mathbb{A}$ and a face image $B$ does not belong to the individual $\mathbb{A}$, The face images $A$ and $B$ are a pair of neighboring face images.
\end{Def}

Then, the sensitivity of facial features is defined as \textbf{Definition \ref{def:sfv}}.

\begin{Def}[The Sensitivity of Facial Features]\label{def:sfv}
If $\mathcal{M}$ is a measurement of face images' facial features, its sensitivity is 
\begin{equation*}
   {\Delta_i} = \mathop{\max}_{A,B}||\mathcal{M}(A)-\mathcal{M}(B)||_1,
\end{equation*}
where $A$ and $B$ are neighboring face images. 
\end{Def}

Finally, the definition of DP is as follows.

\begin{Def}[Differential Privacy]\label{def:dp}
There is a measurement $\mathcal{M}'$ and a pair of neighboring face images $A$ and $B$. $\mathcal{M}'$ is $\varepsilon_0$-differentially private, if the following is satisfied

\begin{equation*}
  \left|\frac{{\Pr \left[  \mathcal{M}'(A)={\bar F}^\prime   \right]}}{{\Pr \left[  \mathcal{M}'(B)={\bar F}^\prime \right]}} \right|\le \exp \left( {{\varepsilon _0}} \right).
\end{equation*}


\end{Def}

\section{Ranked Differential Privacy}
In this section, we introduce our proposed Ranked Differential Privacy (RDP) for facial features with weighted sparsified subspace. Overall, before releasing the original face images, the Laplacian noise is added to the face images' facial features by geometric superposition \cite{9224157,2012Weak}, in order to hide the facial features which are used in the process of face recognition of the systems. 







\subsection{Perturbation in Multiscale Feature Space}\label{sec:wdp}

As well known, the face images' facial features play a key role in the implementation of face recognition. So that they should be protected to avoid adversaries' breaching face recognition-based systems. From this, before the perturbation, facial features should be extracted in the multiscale feature space. Because of its efficiency and rigorous mathematical expression, Multi-level Haar Wavelet Transform (HWT) \cite{2000Multiresolution} is adopted to reduce the dimensionality of the face image's pixels and obtain accurate information about the face image's facial features, \ie, the wavelet coefficient vector constructed by the sparsified wavelet coefficients is obtained. After the perturbation on the wavelet coefficient vectors, the noisy face image could be generated by the noisy pixels matrix ${\bf{P}}'$ obtained by the noisy coefficients vector $\bar C'$ through Multi-level Inverse Wavelet Transform (MIWT).

During the extraction and dimensionality reduction of facial features from face images, we use the first-order Taylor expansion to describe the liner mapping relationship between the noise of the sparse wavelet space and the noisy reduced dimensionality facial features of the noisy face image. After adding Laplacian noise to the elements of the original wavelet coefficient vectors ${\bar C}$, we obtain the noisy feature vector ${{\bar F'}}$ by the noisy face image, which comes from the noisy wavelet coefficient vectors ${\bar C'}$ with MIWT.  According to the first-order Taylor formula for the $i$-th element ${\bar F_i}^\prime $ of the noisy facial feature vector ${{\bar F'}}$, we have
\begin{equation}
\label{eqn:Fi}
\begin{aligned}
{{\bar  F}_i}^\prime & = {{\bar  F}_i} + \sum\limits_{j = 1}^{{M_P}} {\frac{{\partial {{\bar  F}_i}^\prime }}{{\partial {{\bar  C}_j}^\prime }} \cdot \left( {{{\bar  C}_j}^\prime  - {{\bar  C}_j}} \right)}  + o\left( {\bar C'} \right),\\
\end{aligned}
\end{equation}
where $1 \le i \le {M_F}$; ${{\bar C}_j}^\prime $ is the $j$-th element of ${\bar C'}$, where $M_P$ is the number of the elements of the face image's pixel matrix and $1 \le j \le M_P$; and $o\left( {\bar C'} \right)$ is the minimal term, \ie, $o\left( {\bar C'} \right) \sim 0$. Thus, we think
\begin{equation*}
{{\bar F}_i}^\prime  = {{\bar F}_i} + \sum\limits_{j = 1}^{{M_P}} {\frac{{\partial {{\bar F}_i}^\prime }}{{\partial {{\bar C}_j}^\prime }} \cdot \left( {{{\bar C}_j}^\prime  - {{\bar C}_j}} \right)}  .
\end{equation*}
Obviously, ${{{\bar C}_j}^\prime  - {{\bar C}_j}}$ represents the noise of ${{\bar C}_j}^\prime $, and the noise of ${{\bar F}_i}^\prime $ can be considered as the weighted sum of elements' noises of ${\bar C'}$, with that the first-order partial derivative of ${{\bar F}_i}^\prime $ on ${{{\bar C}_j}^\prime }$ is regarded as the weight for the noise of ${{\bar C}_j}^\prime $. 

Then, we define the weight ${w_{ik}}$ for the $i$-th element ${\bar F_i}^\prime $ of the noisy feature vector ${{\bar F'}}$ and the $k$-th element ${\bar C_k}{^ * {}^\prime }$ of the sorted noisy wavelet coefficient vector ${\bar C^ * }{}^\prime $ as ${w_{ik}} = \frac{{\partial {{\bar F}_i}^\prime }}{{\partial {{\bar  C}_k}{{^ * }{}^\prime }}}$. 

After this, with the perturbation provides by weighted sum, we can select the wavelet coefficients more significant in facial features from the sorted wavelet coefficients basis of such weights for adding Laplacian noises with appropriate scale to wavelet coefficients, thereby providing stricter privacy preservation. The definition of the noise mechanism based on such ranked wavelet coefficients is shown in \textbf{Definition} \ref{def:wdp}.

\begin{Def}[Ranked Differential Privacy]\label{def:wdp}
Given the geometric number $K$ between $1$ and ${M_P}$ with mean $\frac{1}{p}$, $0 < p < 1$. For obtaining  the noisy wavelet coefficients vector $\bar C'$, $K$ Laplacian noises with mean $0$ are added to the top $K$ of the wavelet coefficients vector $\bar C$'s ranked elements of a face image, with ranking elements of ${\bar C'}$ by the corresponding weights between the noisy feature vector's elements and such elements. After the inverse transformation with MIWT of $\bar C'$, the noisy pixel matrix ${\bf{P}}'$ is obtained from $\bar C'$, and the noisy face image is generated by ${\bf{P}}'$. Such method,  providing perturbation to the elements of the feature vector ${{\bar F}}$ by the weighted sum of ranked wavelet coefficients' Laplacian noises for protecting facial features, is called Ranked Differential Privacy, referred as "RDP".
\end{Def}



In RDP, after perturbation of wavelet coefficients, the noisy pixels matrix ${\bf{P}}'$ is obtained by ${\bar C^ * }{}^\prime $ through MIWT, and the noisy feature vector of the noisy face image generated by the noisy face image from ${\bf{P}}'$ is $\bar F'$ with ${M_F}$ elements. Let the noise of the $i$-th element ${\bar  F_i}^\prime $ in $\bar F'$ be ${\zeta _i}$, ${{\bar  F}_i}^\prime  = {{\bar  F}_i} + {\zeta _i}$, with  $1 \le i \le {M_F}$. Then we have \textbf{Theorem \ref{theo_4.1}} of RDP about $\bar F'$.

\begin{Theo}\label{theo_4.1}
The noise $\zeta _i$ of the $i$-th element ${\bar  F_i}^\prime $ in $\bar F'$,which is  obtained by RDP, satisfies Laplace distribution, \ie, $\Pr \left[ {{\zeta _i}} \right] = Lap\left( {0,\frac{{{\Delta _i}}}{{{\varepsilon _i}}}} \right)$, where ${{\varepsilon _i}}$ is the privacy budget for the $i$-th element ${\bar  F_i}^\prime $ of $\bar F'$, and $1 \le i \le {M_F}$.
\end{Theo}
\begin{pro}
Let ${{\bar C}}{{^*}^\prime}$ and $\bar C^*$  be the noisy ranked wavelet coefficient vector and the ranked wavelet coefficient vector, separately. According to Eq. (\ref{eqn:Fi}), we have the following first-order Taylor expansion for ${\bar  F_i}^\prime $,
\begin{equation*}
\begin{aligned}
{{\bar F}_i}^\prime = {{\bar  F}_i} + \sum\limits_{k = 1}^{{M_P}} {\frac{{\partial {{\bar  F}_i}^\prime }}{{\partial {{\bar  C}_k}{{^ * }{}^\prime }}} \cdot \left( {{{\bar  C}_k}{{^ * }^\prime } - {{\bar  C}_k}^ * } \right)}  + o\left( {{{\bar C}^ * }{}^\prime } \right),
\end{aligned}
\end{equation*}
where ${{\bar C}_k}{{^*}^\prime}$ and $\bar C_k^*$ are the $k$-th element of ${{\bar C}}{{^*}^\prime}$ and $\bar C^*$, separately, and $o\left({{{\bar C}^ *}{}^\prime} \right)$ are the minimal term, \ie, $o\left( {{{\bar C}^*}{}^\prime } \right) \sim 0$. Thus, we think
\begin{equation*}
{{\bar  F}_i}^\prime  = {{\bar  F}_i} + \sum\limits_{k = 1}^{{M_P}} {\frac{{\partial {{\bar  F}_i}^\prime }}{{\partial {{\bar  C}_k}{{^ * }^\prime }}} \cdot \left( {{{\bar  C}_k}{{^ * }^\prime } - {{\bar  C}_k}^ * } \right)} .
\end{equation*}

The weight for the $i$-th element ${\bar F_i}^\prime $ of the noisy feature vector ${{\bar F'}}$ and the $k$-th element ${\bar C_k}{^ * {}^\prime }$ of the sorted noisy wavelet coefficient vector ${\bar C^ * }{}^\prime $ is ${w_{ik}}$, ${w_{ik}} = \frac{{\partial {{\bar F}_i}^\prime }}{{\partial {{\bar  C}_k}{{^ * }{}^\prime }}}$. $K$ is the given geometric number between $1$ and ${M_P}$ with mean $\frac{1}{p}$, $0 < p < 1$. After we add Laplacian noises to the top $K$ elements of the ranked wavelet coefficient vector ${\bar C^ * } $ according to RDP, the first-order Taylor expansion of ${\bar F_i}^\prime $ is further
\begin{equation*}
   {\bar F_i}^\prime  = {\bar F_i} + \sum\limits_{k = 1}^{{M_P}} {\frac{{\partial {F_i}^\prime }}{{\partial {\bar C_k}{{^ * }^\prime }}}\left( {{\bar C_k}{{^ * }^\prime } - {\bar C_k}^ * } \right)}  = {\bar F_i} + \sum\limits_{k = 1}^{{M_P}} {{w_{ik}}{\xi _k}}  .
\end{equation*}
Then due to there are $K$ noises, ${\xi _k} = \left\{ {\begin{array}{*{20}{c}}
{{\xi _k},\quad 1 \le k \le K}\\
{0,\quad k \le k \le {M_P}}
\end{array}} \right.$, and ${\bar F_i}^\prime  = {\bar F_i} + \sum\limits_{k = 1}^K {{w_{ik}}{\xi _k}} $. From that, we have ${\zeta _i} = \sum\limits_{k = 1}^K {{w_{ik}}{\xi _k}} $, and can get the probability of ${\zeta _i} $, $\Pr \left[ {{\zeta _i}} \right] = \Pr \left[ {\sum\limits_{k = 1}^K {{w_{ik}}{\xi _k}} } \right]$. And based on the geometric superposition, ${\zeta _i} $ satisfies Laplace distribution, \ie, ${\zeta _i} \sim Lap\left( {0,\frac{{{\Delta _i}}}{{{\varepsilon _i}}}} \right)$ with sensitivity ${{\Delta _i}}$ and privacy budget ${{\varepsilon _i}}$ of ${{\bar 
 F_i}^\prime }$. Especially, the scale parameter ${{b_k}}$ of the noise ${\xi _k}$ is related to the privacy budget ${{\varepsilon _i}}$ of ${{\bar 
 F_i}^\prime }$, $1 \le k \le M_P$.

This theorem has been proved.

\end{pro}

Next, the distribution of the noisy feature vector $\bar F'$ can be derived from elements of $\bar F'$. Based on \textbf{Theorem \ref{theo_4.1}}, we can obtain the relationship between the total privacy budget of $\bar F'$ and the scale parameters of Laplacian noises in ${\bar C^ * }{}^\prime $.

\subsection{Privacy Budget Analysis}
In face recognition, the facial feature vector is obtained by projecting the face image onto the eigenfaces, which are constructed by a set of orthogonal eigenvectors through SVD decomposition and dimensionality reduction from the original face images dataset \cite{6793549}. Such eigenvectors capture the most significant variations of features and represent the main change directions in the feature space, and their corresponding singular values represent the importance of eigenfaces. Due to their orthogonality, these eigenvectors are independent and can serve as the basis vectors of the feature space, enabling the representation of the original face image. Consequently, the feature vector's elements, which are projections of the original image onto these eigenvectors, exhibit independence. According to \textbf{Theorem \ref{theo_4.1}}, ${\zeta _i} = \sum\limits_{k = 1}^K {{w_{ik}}{\xi _k}} $, where $K$ is the given geometric number with mean $\frac{1}{p}$, $0 < p < 1$, and ${{w_{ik}}}$ is the weight between ${\bar  F_i}^\prime $ and ${\bar  C_k}{^ * {}^\prime }$, ${w_{ik}} = \frac{{\partial {{\bar F}_i}^\prime }}{{\partial {{\bar  C}_k}{{^ * }{}^\prime }}}$, $1 \le k \le {M_P}$. Let the sensitivity ${{\Delta _i}}$ and the privacy budget ${{\varepsilon _i}}$ of ${\bar F_i}^\prime $, we have $\Pr \left[ {{\zeta _i}} \right] = Lap\left( {0,\frac{{{\Delta _i}}}{{{\varepsilon _i}}}} \right)$. According to the variance operation, the variance of ${\zeta _i}$ is 
\begin{equation*}
   Var\left[ {{\zeta _i}} \right] = 2{\left( {\frac{{{\Delta _i}}}{{{\varepsilon _i}}}} \right)^2}.
\end{equation*}
and the variance of ${\sum\limits_{k = 1}^K {{w_{ik}}{\xi _k}} }$ is
\begin{equation*}
\begin{aligned}
    Var\left[ {\sum\limits_{k = 1}^K {{w_{ik}}{\xi _k}} } \right] &= \sum\limits_{K = 1}^{{M_P}} {p{{\left( {1 - p} \right)}^K}\sum\limits_{k = 1}^K {Var\left[ {{w_{ik}}{\xi _k}} \right]} } \\
& = 2\sum\limits_{K = 1}^{{M_P}} {p{{\left( {1 - p} \right)}^K}\sum\limits_{k = 1}^K {{{\left( {{w_{ik}}{b_k}} \right)}^2}} } \\
& = 2\sum\limits_{k = 1}^{{M_P}} {{{\left( {{w_{ik}}{b_k}} \right)}^2}\sum\limits_{K \ge k}^{{M_P}} {p{{\left( {1 - p} \right)}^K}} } \\
& = 2\sum\limits_{k = 1}^{{M_P}} {{{\left( {{w_{ik}}{b_k}} \right)}^2}\left[ {{{\left( {1 - p} \right)}^k} - {{\left( {1 - p} \right)}^{{M_P} + 1}}} \right]}.
\end{aligned}
\end{equation*}

Because of $Var\left[ {{\zeta _i}} \right] = Var\left[ {\sum\limits_{k = 1}^K {{w_{ik}}{\xi _k}} } \right]$ from $\Pr \left[ {{\zeta _i}} \right] = \Pr \left[ {\sum\limits_{k = 1}^K {{w_{ik}}{\xi _k}} } \right]$, we get the expression of ${\varepsilon _i} $ with respect to ${{b_k}}$ as 
\begin{equation*}
    {\varepsilon _i}\left( {{b_k}} \right) = \frac{{{\Delta _i}}}{{\sqrt {\sum\limits_{k = 1}^{{M_P}} {{{\left( {{w_{ik}}{b_k}} \right)}^2}\left[ {{{\left( {1 - p} \right)}^k} - {{\left( {1 - p} \right)}^{{M_P} + 1}}} \right]} } }}. 
\end{equation*}
And based on that the elements of ${{\bar F'}}$ are independent, the total privacy budget of $\bar F'$ is $\sum\limits_{i = 1}^{{M_F}} {{\varepsilon _i}\left( {{b_k}} \right)} $, then we have the total privacy budget's expression $ \varepsilon \left( {{b_k}} \right) $ of $\bar F'$ with respect to ${{b_k}}$ as
\begin{align}
     \varepsilon \left( {{b_k}} \right) = \sum\limits_{i = 1}^{{M_F}} {\frac{{{\Delta _i}}}{{\sqrt {\sum\limits_{k = 1}^{{M_P}} {{{\left( {{w_{ik}}{b_k}} \right)}^2}\left[ {{{\left( {1 - p} \right)}^k} - {{\left( {1 - p} \right)}^{1+M_P}}} \right]} } }}}.  
     \label{eq:4.1}
\end{align}






\subsection{Optimal Data Utility under Privacy Budget}\label{sec:o}
Our optimization goal is to maximize the noisy face image's visualization quality under the constraint of the privacy budget, \ie, optimizing the noise scale parameters of all wavelet coefficients, which is a constraint optimization problem.

Since the variance of the noise quantifies the negative impact of the noise on the visualization quality and a constant does not have influence on the variance, the cost function for the visualization quality of the face image can be defined by deriving the variance of the noisy pixel matrix, of which value is smaller with the smaller loss of visualization quality. 

Further, the constrained optimization problem can be converted into minimizing the cost function under the constraint of privacy budget. Then, we define the cost function as \textbf{Definition} \ref{def:of}.

\begin{Def}[Cost Function]\label{def:of}
After multi-level HWT for the original face image, ${\bar C}$ is the one-dimensional vector that the wavelet coefficient matrix expansion to. ${{\bar C}^ * }$ is a one-dimensional vector after sorting the elements of ${\bar C}$. According to RDP (Definition \ref{def:wdp}), the Laplacian noises are added  to ${{\bar C}^ * }$ and get the ranked noisy one-dimensional wavelet coefficient vector ${{\bar C}^ * }{}^\prime $. Define that ${{b_k}}$ is the scale parameter of the Laplacian noise of the $k$-th element of  ${{\bar C}^ * }{}^\prime $, $1 \le k \le {M_P}$, then the expression of the cost function $f\left( {{b_k}} \right)$ of the noisy face image with respect to ${{b_k}}$ is 
\begin{align}
    f\left( {{b_k}} \right) = \sum\limits_{k = 1}^{{M_P}} {{{\left( {{b_k}} \right)}^2}\left[ {{{\left( {1 - p} \right)}^k} - {{\left( {1 - p} \right)}^{1+M_P}}} \right]}.  
    \label{eq:4.2}
\end{align}

\end{Def}


\begin{pro}
Given the pixel matrix ${\bf{P}}$ of a face image, and it undergoes a multi-level wavelet transform to obtain a matrix of wavelet coefficients ${\bf{C}}$. Assume that the transformation matrix of wavelet transform is ${\bf{\tilde H}}$ \cite{book, MTAZNAM2018243, BULL2021143}, which is a $\sqrt {{M_P}}  \times \sqrt {{M_P}} $ orthogonal matrix. After $N$-level wavelet transform, let ${\bf{H}} = {{{\bf{\tilde H}}}^N}$, and ${{\bf{H}}^T} = {\left( {{{{\bf{\tilde H}}}^T}} \right)^N}$. Then 
$${\bf{C}} = {{{\bf{\tilde H}}}^N}{\bf{P}} = {\bf{HP}},\ \text{and}\ {\bf{P}} = {\left( {{{\bf{H}}^T}} \right)^N}{\bf{C}} = {{\bf{H}}^T}{\bf{C}},$$
where ${{\bf{H}}^T}{\bf{H}} = {\bf{I}}$, and ${\bf{I}}$ is an identity matrix.

Define the Laplacian noise matrix of wavelet coefficient matrix ${\bf{C}}$ is ${\bf{N}}$, and the noisy wavelet coefficient matrix is ${\bf{C'}} = {\bf{C}} + {\bf{N}}$. Then for the noisy pixel matrix ${{\bf{P'}}}$,
$${\bf{P'}} = {\left( {{{{\bf{\tilde H}}}^T}} \right)^N}{\bf{C'}} = {{\bf{H}}^T}\left( {{\bf{C}} + {\bf{N}}} \right) = {\bf{P}} + {{\bf{H}}^T}{\bf{N}}.$$
Let ${\bf{B}}$ be the Laplacian scale parameter matrix of ${\bf{N}}$, ${{\bf{N}}_{mn}} \sim Lap\left( {0,{{\bf{B}}_{mn}}} \right)$, $1 \le m \le {M_P}$, $1 \le n \le {M_P}$. Thus,
\begin{equation*}
\begin{array}{ll}
Var\left[ {{\bf{P'}}} \right] &  = 2\sum\limits_{m = 1}^{\sqrt {{M_P}} } {\sum\limits_{n = 1}^{\sqrt {{M_P}} } {\sum\limits_{t = 1}^{\sqrt {{M_P}} } {{{\left[ {{{\left( {{{\bf{H}}^T}} \right)}_{mt}}} \right]}^2} \cdot {{\left( {{{\bf{B}}_{tn}}} \right)}^2}} } }  \\
&  = 2\sum\limits_{n = 1}^{\sqrt {{M_P}} } {\sum\limits_{t = 1}^{\sqrt {{M_P}} } {{{\left( {{{\bf{B}}_{tn}}} \right)}^2}\sum\limits_{m = 1}^{{M_P}} {{{\left[ {{{\left( {{{\bf{H}}^T}} \right)}_{mt}}} \right]}^2}} } } \\
& = 2\sum\limits_{n = 1}^{\sqrt {{M_P}} } {\sum\limits_{t = 1}^{\sqrt {{M_P}} } {{{\left( {{{\bf{B}}_{tn}}} \right)}^2}} } \\
& = Var\left[ {{\bf{C'}}} \right] .
\end{array}
\end{equation*}

For the convenience of calculation, we flatten the pixel matrix ${\bf{P}}$ and the wavelet coefficient matrix ${\bf{C}}$ into the one-dimensional pixel vector ${\bar P}$ and the wavelet coefficient vector ${\bar C}$. ${{\bar C}^ * }$ is a one-dimensional vector after sorting the elements of ${\bar C}$. Then the variance of the noisy pixel vector ${\bar P'}$ and the noisy wavelet coefficient vector ${\bar C'}$ is equal, \ie
\begin{equation*}
    \begin{array}{ll}
Var\left[ {\bar P'} \right] &  = Var\left[ {{{\bar C}^ * }{}^\prime } \right]\\
& = 2\sum\limits_{K = 1}^{{M_P}} {p{{\left( {1 - p} \right)}^K}\sum\limits_{k = 1}^K {{{\left( {{b_k}} \right)}^2}} } \\
& = 2\sum\limits_{k = 1}^{{M_P}} {{{\left( {{b_k}} \right)}^2}\left[ {{{\left( {1 - p} \right)}^k} - {{\left( {1 - p} \right)}^{{M_P} + 1}}} \right]} ,
\end{array}
\end{equation*}
where ${{\bar C}^ * }{}^\prime $ is the ranked noisy one-dimensional wavelet coefficient vector.

From that, the expression of the cost function of ${\bar P'}$ with respect to ${{b_k}}$ is as follows 
\begin{equation*}
   f\left( {{b_k}} \right) = \sum\limits_{k = 1}^{{M_P}} {{{\left( {{b_k}} \right)}^2}\left[ {{{\left( {1 - p} \right)}^k} - {{\left( {1 - p} \right)}^{{M_P} + 1}}} \right]}   ,
\end{equation*}
which is the cost function of the noisy face image.

\end{pro}


From that, a constant $\varepsilon_0$ is given as the total privacy budget of the noisy feature vector $\bar F'$, and the constraint is $\varepsilon \left( {{b_k}} \right) = \varepsilon_0$. Thus, the optimization problem is as follows,
\begin{equation*}
  {b_k} = \mathop {{\rm{argmin}}}\limits_{{b_k}} \left\{ {f\left( {{b_k}} \right)|\varepsilon \left( {{b_k}} \right) = {\varepsilon _0}} \right\}.
\end{equation*}

We solve this optimization problem according to the Lagrangian Multiplier method (LM) to better balance the need to sanitize the facial features and preserve the visualization quality of the face image. According to LM, the Lagrangian function with respect to ${b _k}$ is
\begin{equation*}
    L\left( {{b_k},\lambda } \right) = f\left( {{b_k}} \right) + \lambda \left( {\varepsilon \left( {{b_k}} \right) - {\varepsilon _0}} \right),
\end{equation*}
which can be solved by setting the first derivatives with respect to $\lambda$ and $b_k$ to zeros,
$$\left\{\begin{array}{lr}
{\frac{{\partial L\left( {b_k} \right)}}{{\partial \lambda }} = 0}; \nonumber \\
{\frac{{\partial L\left( {b_k} \right)}}{{\partial {b_k}}} = 0}, \nonumber
\end{array}\right.$$
where the first equation reduces to the following,
\begin{equation} 
\sum\limits_{i = 1}^{{M_F}} {\frac{{\lambda {\Delta_i}{{\left( {{w_{ik}}} \right)}^2}}}{{{{\left\{ {\sum\limits_{k = 1}^{{M_P}} {{{\left( {{w_{ik}}{b_k}} \right)}^2}\left[ {{{\left(1-p\right)}^k}-{{\left(1-p \right)}^{1+M_P}}} \right]} } \right\}}^{\frac{3}{2}}}}}} = 2. 
\label{eq:4.3}
\end{equation}

It is clear that Eq. (\ref{eq:4.3}) is nonlinear. Here we propose two ways to solve the nonlinear LM  equations for two different applications, \ie, 1) the Normalization Approximation (NA) for the real-time online applications; and 2) the LM optimization Gradient Descent (LMGD) for the offline applications that need more accurate result.

\subsubsection{Optimization with Normalization Approximation}\label{sec:na}
It is clear that Eq. (\ref{eq:4.3}) is nonlinear,  which makes it difficult to find the analytical solution. Therefore, we propose the optimization with Normalization Approximation (NA), which assumes that all wavelet coefficients of interest contribute equally to the total privacy budget $\varepsilon_0$. 

To achieve such goal, it would be intuitive to choose the weighted Laplacian noise scale parameters ${b_k}$ of Eq. (\ref{eq:4.3}) to be the average value, \ie, ${w_{ik}}{b_k} \approx  \bar b = \frac{{\sum\limits_{i = 1}^{{M_F}} {{w_{ik}}{b_k}} }}{{{M_F}}}$, where $1 \le i \le {M_F}$ and $1 \le k \le {M_P}$. And substituting ${\bar b}$ into Eq. (\ref{eq:4.3}), we get
\begin{equation*}
    \sum\limits_{i = 1}^{{M_F}} {\frac{{\lambda {\Delta _i}{{\left( {{w_{ik}}} \right)}^2}}}{{{{\bar b}^3}{{\left\{ {\sum\limits_{k = 1}^{{M_P}} {{{\left( {{w_{ik}}} \right)}^2}\left[ {{{\left( {1 - p} \right)}^k} - {{\left( {1 - p} \right)}^{1 + {M_P}}}} \right]} } \right\}}^{\frac{3}{2}}}}}}  = 2.
\end{equation*}
It can be substituted by ${w_{ik}}{b_k}$ to replace ${\bar b}$ for obtaining the equations about every ${b_k}$ and $\lambda $, which can be used to solve $\frac{{\partial L\left( {b_k} \right)}}{{\partial {b_k}}} = 0$. And then, Eq. (\ref{eq:4.3}) can be solved as follows,
\begin{equation*}
{b_k} = \frac{{\sum\limits_{i = 1}^{{M_F}} {\frac{{{\Delta }_i}}{{{w_{ik}}}}} }}{{{\varepsilon _0}{{\left[{\frac{{1 - p - {{\left( {1 - p} \right)}^{1+M_P}}}}{p} - {M_P}{{\left( {1 - p} \right)}^{1+M_P}}} \right]}^{\frac{1}{2}}}}}.
\label{eq:4.4}
\end{equation*}

\subsubsection{LM Optimization via Gradient Descent}\label{sec:crc}
Although NA can solve the system of nonlinear equations in this optimization problem in real time, the assumption as its basis is are very idealistic, which may cause errors far from the best result. For the case allowing solving this problem offline, we propose solving the LM optimization problem via Gradient Descent (LMGD), which continuously tunes the initial value of $b_k$ until an optimal solution is obtained through iterative gradient descent updating. 

It aims to find the best solution for $b_k$ and $\lambda$ by solving following equations,
$$\left\{
\begin{array}{lr}
{\frac{{\partial L\left( {{b_k}} \right)}}{{\partial \lambda }} = 0};\\
{\frac{{\partial L\left( {{b_k}} \right)}}{{\partial {b_k}}} = 0},
\end{array}
\right.$$ which can be solved with two alternating gradient descent updates: 

1) gradient descent update with respect to $b_k$; 

2) gradient descent update with respect to $\lambda$. 

However, simple alternating gradient descent updates have the problem of finding the optimal step length or learning rate, which is difficult for non-convex optimization problems. Also, the nonlinear operation is required to achieve the minimum $b_k$, which motivates us to develop the LMGD to solve the nonlinear LM problem.  

According to LM, the purpose of LMGD is to obtain the optimal solution of $b_k$ and $\lambda$,  through solving the following equations, 
\begin{equation*}
   \left\{ {\begin{array}{*{20}{c}}
{\sum\limits_{i = 1}^{{M_F}} {\frac{{{\Delta _i}}}{{\sqrt {\sum\limits_{k = 1}^{{M_P}} {{{\left( {{w_{ik}}{b_k}} \right)}^2}\left[ {{{\left( {1 - p} \right)}^k} - {{\left( {1 - p} \right)}^{1 + {M_P}}}} \right]} } }}}  - {\varepsilon _0} = 0,}\\
{\sum\limits_{i = 1}^{{M_F}} {\frac{{\lambda {\Delta _i}{{\left( {{w_{ik}}} \right)}^2}}}{{{{\left\{ {\sum\limits_{k = 1}^{{M_P}} {{{\left( {{w_{ik}}{b_k}} \right)}^2}\left[ {{{\left( {1 - p} \right)}^k} - {{\left( {1 - p} \right)}^{1 + {M_P}}}} \right]} } \right\}}^{\frac{3}{2}}}}}}  - 2 = 0.}
\end{array}} \right.
\end{equation*}
We define the error functions of two updating sub-tasks in LMGD as follows, 
\begin{align*}
\begin{aligned}
& {F_1}\left( {b_k^{\left( h \right)}} \right)\\
 = & \left| {\sum\limits_{i = 1}^{{M_F}} {\frac{{{\Delta _i}}}{{\sqrt {\sum\limits_{k = 1}^{{M_P}} {{{\left( {{w_{ik}}b_k^{\left( h \right)}} \right)}^2}\left[ {{{\left( {1 - p} \right)}^k} - {{\left( {1 - p} \right)}^{1 + {M_P}}}} \right]} } }}}  - {\varepsilon _0}} \right|;\\
 \quad \\
& {F_2}\left( {b_k^{\left( h \right)},{\lambda ^{\left( h \right)}}} \right)\\
 = & \left| {\sum\limits_{i = 1}^{{M_F}} {\frac{{{\lambda ^{\left( h \right)}}{\Delta _i}{{\left( {{w_{ik}}} \right)}^2}}}{{{{\left\{ {\sum\limits_{k = 1}^{{M_P}} {{{\left( {{w_{ik}}b_k^{\left( h \right)}} \right)}^2}\left[ {{{\left( {1 - p} \right)}^k} - {{\left( {1 - p} \right)}^{1 + {M_P}}}} \right]} } \right\}}^{\frac{3}{2}}}}}}  - 2} \right|,
\end{aligned}
\end{align*}
where $h$ represents the $h$-th parameter update. From that, the total loss function is
\begin{equation*}
F\left( {b_k^{\left( h \right)},{\lambda ^{\left( h \right)}}} \right) = {F_1}\left( {b_k^{\left( h \right)}} \right) + {F_2}\left( {b_k^{\left( h \right)},{\lambda ^{\left( h \right)}}} \right).
\end{equation*}

When ${F_1}\left( {b_k^{(h)}} \right) \to 0$ with ${F_2}\left( {b_k^{\left( h \right)},{\lambda^{\left( h \right)}}} \right) \to 0$, the closer $b_k$ and $\lambda$ are to the optimal solution. 

The LMGD takes an initial solution $b_k^{\left( 0 \right)}$ of $b_k$ and ${{\lambda ^{\left( 0 \right)}}}$ of $\lambda$ as input, where $\varepsilon \left( {b_k^{\left( 0 \right)}} \right) \approx {\varepsilon _0}$, however $\varepsilon \left( {b_k^{\left( 0 \right)}} \right) \ne {\varepsilon _0}$. After the initial values are inputted, the optimal solution of $b_k$ and $\lambda$ is obtained through the iterative updating process updated by two alternating gradient descent until the loss function $F\left( {b_k^{\left( h \right)},{\lambda ^{\left( h \right)}}} \right)$ is less than a minimum value $\delta $, which is very close to $0$. 

Given the step length or learning rate of $\eta $. Now the iterative updating process takes two alternating updates as follows,

(1) Step 1: firstly, $b_k^{(h)}$ is updated iteratively by minimizing the loss function ${F_1}\left( {b_k^{\left( h \right)}} \right)$ using the gradient descent, while $b_k^{(h)}$ and $\lambda ^{(h)}$ are updated iteratively by minimizing the loss function ${F_2}\left( {b_k^{(h)},{\lambda ^{\left( h \right)}}} \right)$ using gradient descent, \ie,
\begin{equation*}
\begin{array}{ll}
    &b_k^{\left( h \right)} = b_k^{\left( h \right)} - \eta \left[ {\frac{{\partial {F_1}\left( {b_k^{\left( h \right)}} \right)}}{{\partial b_k^{\left( h \right)}}} + \frac{{\partial {F_2}\left( {b_k^{\left( h \right)},{\lambda ^{\left( h \right)}}} \right)}}{{\partial b_k^{\left( h \right)}}}} \right];\\

    &{\lambda ^{\left( h \right)}} = {\lambda ^{\left( h \right)}} - \eta \frac{{\partial {F_2}\left( {b_k^{\left( h \right)},{\lambda ^{\left( h \right)}}} \right)}}{{\partial {\lambda ^{\left( h \right)}}}},
\end{array}
\end{equation*}
where the gradient ${\frac{{\partial {F_1}\left( {b_k^{\left( h \right)}} \right)}}{{\partial b_k^{\left( h \right)}}}}$ can be obtained as follows
\begin{align}
\begin{array}{ll}
    {\frac{{\partial {F_1}\left( {b_k^{\left( h \right)}} \right)}}{{\partial b_k^{\left( h \right)}}}} = \left| {\sum\limits_{i = 1}^{{M_F}} {\frac{{{\Delta _i}b_k^{\left( h \right)}{{\left( {{w_{ik}}} \right)}^2}\left[ {{{\left( {1 - p} \right)}^k} - {{\left( {1 - p} \right)}^{1 + {M_P}}}} \right]}}{{{{\left\{ {\sum\limits_{k = 1}^{{M_P}} {{{\left( {{w_{ik}}b_k^{\left( h \right)}} \right)}^2}\left[ {{{\left( {1 - p} \right)}^k} - {{\left( {1 - p} \right)}^{1 + {M_P}}}} \right]} } \right\}}^{\frac{3}{2}}}}}} } \right|.
     \label{eq:4.5}
\end{array}
\end{align}

(2) Step 2: then, the gradients of ${\frac{{\partial {F_2}\left( {b_k^{\left( h \right)}, {\lambda ^{\left( h \right)}}} \right)}}{{\partial b_k^{\left( h \right)}}}}$ and $\frac{{\partial {F_2}\left( {b_k^{\left( h \right)},{\lambda ^{\left( h \right)}}} \right)}}{{\partial {\lambda ^{\left( h \right)}}}}$ are calculated by
\begin{equation*}
\begin{aligned}
    &\frac{{\partial {F_2}\left( {b_k^{\left( h \right)},{\lambda ^{\left( h \right)}}} \right)}}{{\partial b_k^{\left( h \right)}}}\\
 = & \left| {3\sum\limits_{i = 1}^{{M_F}} {\frac{{{\lambda ^{\left( h \right)}}b_k^{\left( h \right)}{\Delta _i}{{\left( {{w_{ik}}} \right)}^4}\left[ {{{\left( {1 - p} \right)}^k} - {{\left( {1 - p} \right)}^{1 + {M_P}}}} \right]}}{{{{\left\{ {\sum\limits_{k = 1}^{{M_P}} {{{\left( {{w_{ik}}b_k^{\left( h \right)}} \right)}^2}\left[ {{{\left( {1 - p} \right)}^k} - {{\left( {1 - p} \right)}^{1 + {M_P}}}} \right]} } \right\}}^{\frac{5}{2}}}}}} } \right|;\\
& \frac{{\partial {F_2}\left( {b_k^{\left( h \right)},{\lambda ^{\left( h \right)}}} \right)}}{{\partial {\lambda ^{\left( h \right)}}}}\\
 = & \left| {\sum\limits_{i = 1}^{{M_F}} {\frac{{{\Delta _i}{{\left( {{w_{ik}}} \right)}^2}}}{{{{\left\{ {\sum\limits_{k = 1}^{{M_P}} {{{\left( {{w_{ik}}b_k^{\left( h \right)}} \right)}^2}\left[ {{{\left( {1 - p} \right)}^k} - {{\left( {1 - p} \right)}^{1 + {M_P}}}} \right]} } \right\}}^{\frac{3}{2}}}}}} } \right|.
\end{aligned}
\end{equation*}

(3) Step 3: finally, after alternatively repeating Step 1 and Step 2 above, the optimal solution of $b_k$ and $\lambda$ are obtained until $F\left( {b_k^{\left( h \right)},{\lambda ^{\left( h \right)}}} \right) < \delta $. 

The process by Eq. (\ref{eq:4.5}) is proved as follow:

\begin{pro}
If the noise's scale parameter of the $i$-th element ${{F}_i}^\prime $ of the noisy feature vector $\bar F'$ is ${b_i}^{\left( {\bar F} \right)}$, then the expression of ${b_i}^{\left( {\bar F} \right)}$ with respect to ${{b^{\left( h \right)}}_k}$ is
\begin{equation*}
   b_i^{\left( {\bar F} \right)}\left( {b_k^{\left( h \right)}} \right) = \sqrt {\sum\limits_{k = 1}^{{M_P}} {{{\left( {{w_{ik}} b_k^{\left( h \right)}} \right)}^2}\left[ {{{\left( {1 - p} \right)}^k} - {{\left( {1 - p} \right)}^{{M_P} + 1}}} \right]} }  , 
\end{equation*}
and the expression of the total privacy budget $\varepsilon$ with respect to ${{b^{\left( h \right)}}_k}$ is 
\begin{equation*}
 \varepsilon \left( {b_k^{\left( h \right)}} \right) = \sum\limits_{i = 1}^{{M_F}} {\frac{{{\Delta _i}}}{{\sqrt {\sum\limits_{k = 1}^{{M_P}} {{{\left( {{w_{ik}} b_k^{\left( h \right)}} \right)}^2}\left[ {{{\left( {1 - p} \right)}^k} - {{\left( {1 - p} \right)}^{{M_P} + 1}}} \right]} } }}}  ,
\end{equation*}
where  ${\Delta }_i$ is the sensitivity of the $i$-th element ${{\bar F}_i}^\prime $ of $\bar F'$, $1 \le i \le {M_F}$, ${{w_{ik}}}$ is the weights between ${{\bar F}_i}^\prime $ and ${{\bar  C}_k}{^ * {}^\prime }$,  $1 \le k \le {M_P}$, and $p$ is the probability parameter greater than $0$ and less than $1$.

According to the chain rule, 
\begin{equation*}
 \frac{{\partial \varepsilon \left( {b_k^{\left( h \right)}} \right)}}{{\partial b_k^{\left( h \right)}}} = \frac{{\partial \varepsilon \left( {b_k^{\left( h \right)}} \right)}}{{\partial b_i^{\left( {\bar F} \right)}\left( {b_k^{\left( h \right)}} \right)}} \cdot \frac{{\partial b_i^{\left( {\bar F} \right)}\left( {b_k^{\left( h \right)}} \right)}}{{\partial b_k^{\left( h \right)}}} .
\end{equation*}
 Due to the independence of the elements of the feature vector,
\begin{equation*}
   \frac{{\partial \varepsilon \left( {b_k^{\left( h \right)}} \right)}}{{\partial b_i^{\left( {\bar F} \right)}}} =  - \frac{{{\Delta _i}}}{{{{\left( {b_i^{\left( {\bar F} \right)}} \right)}^2}}},
\end{equation*}
and
\begin{equation*}
 \frac{{\partial b_i^{\left( {\bar F} \right)}}}{{\partial b_k^{\left( h \right)}}} = \frac{{b_k^{\left( h \right)}{{\left( {{w_{ik}}} \right)}^2}\left[ {{{\left( {1 - p} \right)}^k} - {{\left( {1 - p} \right)}^{{M_P} + 1}}} \right]}}{{\sqrt {\sum\limits_{k = 1}^{{M_P}} {{{\left( {{w_{ik}} b_k^{\left( h \right)}} \right)}^2}\left[ {{{\left( {1 - p} \right)}^k} - {{\left( {1 - p} \right)}^{{M_P} + 1}}} \right]} } }}.
\end{equation*}

Next, we substitute the above expression into the derivative $ \frac{{\partial \varepsilon \left( {b_k^{\left( h \right)}} \right)}}{{\partial b_k^{\left( h \right)}}}$, and obtain it as follows
\begin{equation*}
\begin{aligned}
& \frac{{\partial {F_1}\left( {b_k^{\left( h \right)}} \right)}}{{\partial b_k^{\left( h \right)}}}\\
= & \left| {\sum\limits_{i = 1}^{{M_F}} {\frac{{{\Delta _i}b_k^{\left( h \right)}{{\left( {{w_{ik}}} \right)}^2}\left[ {{{\left( {1 - p} \right)}^k} - {{\left( {1 - p} \right)}^{{M_P} + 1}}} \right]}}{{{{\left\{ {\sum\limits_{k = 1}^{{M_P}} {{{\left( {{w_{ik}}b_k^{\left( h \right)}} \right)}^2}\left[ {{{\left( {1 - p} \right)}^k} - {{\left( {1 - p} \right)}^{{M_P} + 1}}} \right]} } \right\}}^{\frac{3}{2}}}}}} } \right|.
\end{aligned}
\end{equation*}    
  
\end{pro}

\section{Analysis of Privacy and Utility}


\textbf{Privacy.} Now let's look at the privacy settings of the RDP for the feature vectors ${\bar F'}$ of the noisy face image according to \textbf{Definition} \ref{def:wdp}. In this paper, the elements of the feature vector are independent of each other with the given total privacy budget $\varepsilon _0$. 

\begin{Theo}\label{theo_5.1}
RDP satisfies $\varepsilon _0$-differential privacy.

\end{Theo}

\begin{pro}
The face images $A$ and $B$ are a pair of neighboring face images. According to \textbf{Definition} \ref{def:dp}, proving that RDP satisfies $\varepsilon _0$-differentially privacy is equivalent to prove 
\begin{equation*}
\left| {\frac{{\Pr \left[ {{{\bar F}^{\left( A \right)}}{}^\prime } \right]}}{{\Pr \left[ {{{\bar F}^{\left( B \right)}}{}^\prime } \right]}}} \right| \le \exp \left( {{\varepsilon _0}} \right),
\end{equation*}
where ${{{\bar F}^{\left( {A} \right)}}{}^\prime }$ and ${{\bar F}^{\left( B \right)}}{}^\prime$ is the noisy feature vectors of $A$ and $B$.

Let ${\bar F^{\left( \bar A \right)}}$ be the standard feature vector of a certain individual $\mathbb{A}$'s standard face ${\bar A}$. Then, ${{\Delta _i}}$ is the sensitivity fot the $i$-th element of the noisy feature vector, and the Laplace scale parameter of that is $b_i^{\left( {\bar F} \right)}$. Define the privacy budget forthe $i$-th element of the noisy feature vector as $\varepsilon _i$, ${\varepsilon _i} = \frac{{{\Delta _i}}}{{b_i^{\left( {\bar F} \right)}}}$, $1 \le i \le {M_F}$. Let ${\varepsilon _0} = \sum\limits_{i = 1}^{{M_F}} {{\varepsilon_i}} $, 
\begin{equation*}
 \begin{aligned}
\left| {\frac{{\Pr \left[ {{{\bar F}^{\left( A \right)}}{}^\prime } \right]}}{{\Pr \left[ {{{\bar F}^{\left( B \right)}}{}^\prime } \right]}}} \right| = & \prod\limits_{i = 1}^{{M_F}} {\frac{{\left| {\Pr \left[ {{{\bar F}_i}{{^{\left( A \right)}}{}^\prime }} \right]} \right|}}{{\left| {\Pr \left[ {{{\bar F}_i}{{^{\left( B \right)}}{}^\prime }} \right]} \right|}}} \\
 = & \prod\limits_{i = 1}^{{M_F}} {\frac{{\left| {\exp \left( { - \frac{{\left| {{{\bar F}_i}{{^{\left( A \right)}}{}^\prime } - {{\bar F}_i}^{\left( {\bar A} \right)}} \right|}}{{{b_i}^{\left( {\bar F} \right)}}}} \right)} \right|}}{{\left| {\exp \left( { - \frac{{\left| {{{\bar F}_i}{{^{\left( B \right)}}{}^\prime } - {{\bar F}_i}^{\left( {\bar A} \right)}} \right|}}{{{b_i}^{\left( {\bar F} \right)}}}} \right)} \right|}}} \\
 = & \prod\limits_{i = 1}^{{M_F}} {\left| {\exp \left( {\frac{{\left| {{{\bar F}_i}{{^{\left( B \right)}}{}^\prime } - {{\bar F}_i}^{\left( {\bar A} \right)}} \right| - \left| {{{\bar F}_i}{{^{\left( A \right)}}{}^\prime } - {{\bar F}_i}^{\left( {\bar A} \right)}} \right|}}{{{b_i}^{\left( {\bar F} \right)}}}} \right)} \right|} \\
 \le & \left| {\prod\limits_{i = 1}^{{M_F}} {\exp \left( {\frac{{\left| {{{\bar F}_i}{{^{\left( B \right)}}{}^\prime } - {{\bar F}_i}{{^{\left( A \right)}}{}^\prime }} \right|}}{{{b_i}^{\left( {\bar F} \right)}}}} \right)} } \right|\\
 \le & \left| {\prod\limits_{i = 1}^{{M_F}} {\exp \left( {\frac{{{\Delta _i}}}{{{b_i}^{\left( {\bar F} \right)}}}} \right)} } \right|\\
 = & \exp \left( {{\varepsilon _0}} \right)
\end{aligned}
\end{equation*}

Now we have
\begin{equation*}
\left| {\frac{{\Pr \left[ {{{\bar F}^{\left( A \right)}}{}^\prime } \right]}}{{\Pr \left[ {{{\bar F}^{\left( B \right)}}{}^\prime } \right]}}} \right| \le \exp \left( {{\varepsilon _0}} \right),
\end{equation*}
from which \textbf{Theorem} \ref{theo_5.1} is proved.
\end{pro}


\textbf{Data utility.} In this paper, we use the face image's PSNR (Peak Signal-to-Noise Ratio) value to represent the data utility (\ie\ visualization quality) of the face image. PSNR is a measure to evaluate image quality. The higher the value, the less the face image distortion. PSNR of the noisy face image is
\begin{equation}
\label{eq:PSNR}
   PSNR = 10 \cdot {\log _{10}}\left( {\frac{{{{\left( {\max \left\{ {\bar P'} \right\}} \right)}^2}}}{{{\sigma _r}^2}}} \right),
\end{equation}
where ${\bar P'}$ is the one-dimensional pixel vector after the noisy pixel matrix ${\bf{P}'}$ flattened, ${\max \left\{ {\bar P'} \right\}}$ is the maximum element of ${\bar P'}$, and ${{\sigma _r}^2}$ is the real variance of ${\bar P'}$. The higher PSNR is, the more image quality of the noisy face image retains, \ie, the smaller the difference with the original face image.

\section{The Optimized RDP Algorithm}

Algorithm \ref{alg1} summarizes the optimized RDP: the inputs include the original face image pixel matrix ${\bf{P}}$, the total privacy budget  $\varepsilon _0$, and the geometric superposition parameter $p \in [0, 1]$; and the outputs are the noise sanitized face image pixel matrix ${\bf{P'}}$ and the noisy feature vector ${\bar F'}$.

First, we obtain the original sparse feature coefficient matrix ${\bf{C}}$ by multiscale feature dimensionality reduction of the original face image, which can be cast into the one-dimensional feature coefficient vector ${\bar C}$. Then, the influence or weight ${w_{ik}}$ the $k$-th element ${{\bar C}_k}{^ * {}^\prime }$ of the sorted noisy wavelet coefficient vector ${{\bar C}^ * }{}^\prime  $, with respect to the $i$-th element ${\bar F}_i^{\prime}$ of the noisy feature vector ${{\bar F'}}$, is calculated, with $1 \le i \le {M_F}$, $1 \le k \le {M_P}$. After that, we obtain the optimized scale parameter $b_k$ of the Laplacian noise for the $k$-th element ${{\bar C}_k}{^ * {}^\prime }$ of the sorted noisy feature coefficient vector ${{\bar C}^ * }{}^\prime  $ according to the NA or the LMGD, with $1 \le k \le {M_P}$. Finally, the sorted noisy feature coefficient vector ${{\bar C}^ * }{}^\prime  $ is obtained through the \textbf{Definition} \ref{def:wdp}, from which we obtain the noisy feature coefficient matrix ${{\bf{C'}}}$ and the noisy feature vector ${\bar F'}$ of the noisy face image generated from ${{\bf{C'}}}$.

\begin{algorithm}[htbp]
\caption{The Optimized RDP Algorithm}\label{alg1}
\begin{algorithmic}[1]
\Ensure The original face image pixel matrix ${\bf{P}}$, the total privacy budget $\varepsilon _0$, the probability parameters $p$
\Require The noisy face image pixel matrix ${\bf{P'}}$, the noisy feature vector ${\bar F'}$
\State Obtain ${{\bf{C}}}$ from ${\bf{P}}$ through multiscale feature dimensionality reduction;
\State ${{\bf{C}}}$ is cast into ${\bar C}$; 
\State sort ${\bar C}$ to get ${{\bar C}^ * }  $ according to the amplitudes of its elements;
\For {$i = 1 \to {M_F}$} 
\For {$k = 1 \to {M_P}$} 
\State ${w_{ik}} = \frac{{\partial {{\bar F}_i}^\prime }}{{\partial {{\bar C}_k}{{^ * }{}^\prime }}}$;
\EndFor
\EndFor
\For {$k = 1 \to {M_P}$} 
\State Calculate  $b_k$ according to the NA or the LMGD;
\EndFor
\State Sample the geometric number $K \sim Geo\left( p \right)$;
\State Initialize ${{\bar C}^ * }{}^\prime  $;
\For {$k = 1 \to {M_P}$} 
\If {$k \le K$}
\State Sample the Laplacian noise ${\xi _k} \sim Lap\left( {0,{b_k}} \right)$;
\State ${\bar C_k}{^ *{} ^\prime } = {\bar C_k}^ *  + {\xi _k}$;
\Else
\State ${\bar C_k}{^ *{} ^\prime } = {\bar C_k}^ * $;
\EndIf
\EndFor
\State ${{\bar C}^ * }{}^\prime  $ is cast into ${{\bf{C'}}}$;
\State Obtain ${{\bf{P'}}}$ from ${{\bf{C'}}}$ through MIWT;
\State Obtain ${\bar F'}$ according to ${\bf{P'}}$;
\State \Return{${\bf{P'}}$, ${\bar F'}$};
\end{algorithmic}
\end{algorithm}

Now let's look at the time complexity and storage complexity of Algorithm \ref{alg1}, which is mainly determined by two steps: 1) the step of calculating the weights between the noisy feature vector ${{\bar F'}}$ and the sorted noisy feature coefficient vector ${{\bar C}^ * }{}^\prime  $; and 2) the step of obtaining the optimized Laplacian noise scale parameters of ${{\bar C}^ * }{}^\prime  $.

Step 1) calculating the weights between the noisy feature vector ${{\bar F'}}$ and the sorted noisy feature coefficient vector ${{\bar C}^ * }{}^\prime  $ (lines 4-8): because ${{\bar F'}}$ has ${M_F}$ element and ${{\bar C}^ * }{}^\prime  $ has ${M_P}$ elements, the number of weights is ${M_F} \times {M_P}$. Then the time complexity is ${\mathcal{O}}\left( {{M_F}{M_P}} \right)$ and the storage complexity of this stage is ${\mathcal{O}}\left( {{M_F}{M_P}} \right)$. 

Step 2) obtaining the optimized Laplacian noise scale parameters of ${{\bar C}^ * }{}^\prime  $ (lines 9-11): the number of elements in ${{\bar C}^ * }{}^\prime  $ is ${M_P}$. When the RDP is optimized by NA,  the temporal complexity is ${\mathcal{O}}\left( 1 \right)$ and the spatial complexity is ${\mathcal{O}}\left( {{M_P}} \right)$. When the optimization is LMGD, the time complexity is ${\mathcal{O}}\left( {{M_P}} \right)$ and the spatial complexity is ${\mathcal{O}}\left( {{M_P}} \right)$. 
  
Since these two steps are sequentially composed, the time complexity is ${\mathcal{O}}\left( {{M_F}{M_P}} \right)$ and the spatial complexity is ${\mathcal{O}}\left( {{M_F}{M_P} + {M_P}} \right)$ if the scale parameters are optimized by NA; and the time complexity is $\mathcal{O}\left( {M_F}{M_P} + {M_P} \right)$ and the spatial complexity is ${\mathcal{O}}\left( {{M_F}{M_P} + {M_P}} \right)$ if the scale parameters are optimized by the LMGD. 
\section{Experimental Evaluations}

We implement our simulations with Python 3.10 on a laptop with Intel Core i7-13700KF, 3.40GHz and Windows 10 system equipped with 64GB main memory.

\subsection{Experimental Setup}

\subsubsection{Datasets}

We use the following datasets to conduct experiments:
\par
\textbf{Labeled Faces in the Wild (LFW)} \cite{huang:inria-00321923} contains more than $13000$ images of $5749$ identities and provides a standard benchmark for face verification, composed of $6000$ face pairs with $3000$ matched and the other $3000$ non-matched.
\par
\textbf{PubFig83} \cite{5981788} contains $8300$ cropped face images, made up of $100$ images for each of $83$ public figures.
\subsubsection{Evaluation Metrics}

We use the following metrics in the experiment for performance comparisons: 

\textbf{Variance}. This metric measures the visualization quality of the noise sanitized face images. Because noise will have a negative impact on the quality of the face image, the smaller the variance of the face image's noise, the higher the retained quality of the face image. In this paper, we compare the theoretical variance and the true variance of the noisy image of five methods over the given range from $\varepsilon_0 = 0.2$ to $\varepsilon_0 = 1.0$ of the total privacy budget with probability parameter $p=0.02$. Since the noise follows the Laplace distribution, the theoretical variance ${\sigma _t}^2= 2\sum\limits_{k = 1}^K {{{\left( {{b_k}} \right)}^2}}$, 
where ${{b_k}}$ is the scale parameter of Laplacian noise of the $k$-th element ${{\bar  C}_k}{^ * {}^\prime }$ of the sorted feature coefficient vector ${{\bar C}^ * }{}^\prime  $ and $K$ is the geometric number with mean $\frac{1}{p}$.  

The real or actual variance during the numerical experiment, denoted as $ {\sigma _r}^2$, is given by ${\sigma _r}^2 = \sum\limits_{k = 1}^{{M_P}} {{{\left( {{{\bar  P}_k}^\prime  - {{\bar  P}_k}} \right)}^2}}$,
where ${{\bar P_k}^\prime }$ is the $k$-th element of the noisy one-dimensional pixel vector ${\bar P'}$ of the face image, and ${{\bar P_k}}$ is the $k$-th element of the original one-dimensional pixel vector ${\bar P}$ of the face image.

\textbf{Structural Similarity Index (SSIM)} \cite{1284395}. We conduct SSIM between the noise sanitized face images and the original face images. It imitates the human visual system and quantifies the attributes of images from brightness, contrast and structure. It uses the mean to estimate brightness, the variance to estimate contrast, and the covariance to estimate structural similarity.

\textbf{Peak Signal-to-Noise Ratio (PSNR)}. This metric evaluates the visualization quality of the sanitized face images. It is calculated by Eq. (\ref{eq:PSNR}). The higher the PSNR value, the better the visualization quality of the sanitized face images or the less the face image distortion. 

\textbf{False Negative Rate (FNR)}. This metric measures the effect of protection against the bypassing attack. It refers to the ratio of face images that should be identified as this individual but not. The larger the FNR is, the smaller the probability that the real facial features in this face image can be obtained, and the stronger protection for the method provides. 

In particular, for the accuracy of face recognition after adding noise, ${\Omega ^{\left( \mathbb{A} \right)}}$ represents the face image dataset of the individual $\mathbb{A}$, and ${F_\mathbb{A}}$ is a function that outputs the face recognition results, representing the feature vector of any face image $A$ and the standard of $\mathbb{A}$ whether the feature vector of the face satisfies Eq. (\ref{eq:3.1}). If it does, the output is $1$, \ie, ${{F_\mathbb{A}}\left( A \right) = 1}$, indicating that the result of face recognition is correct, otherwise the output is $0$. $f'$ represents the differential privacy mechanism with noise. Then, before adding noise, the accuracy is 1, that is $\Pr \left[ {\left. {{F_\mathbb{A}}\left( A \right) = 1} \right|A \in {\Omega ^{\left( \mathbb{A} \right)}}} \right] = 1$. Let $\Pr \left[ {\left. {{F_A}\left( {f'\left( A \right)} \right) = 0} \right|A \in {\Omega ^{\left( A \right)}}} \right] = {\hat p}$, then 
\begin{equation*}
    \begin{aligned}
    \hat p = \prod\limits_{i = 1}^{{M_F}} {\frac{{{\varepsilon _i}}}{{2{\Delta _i}}}}  \cdot \exp \left( { - \sum\limits_{i = 1}^{{M_F}} {{\varepsilon _i}} } \right) = &\prod\limits_{i = 1}^{{M_F}} {\frac{{{\varepsilon _i}}}{{2{\Delta _i}}}}  \cdot \exp \left( { - {\varepsilon _0}} \right)\\
 \le &{\left( {\frac{{{\varepsilon _0}}}{2}} \right)^{{M_F}}}\frac{{\exp \left( { - {\varepsilon _0}} \right)}}{{\prod\limits_{i = 1}^{{M_F}} {{\Delta _i}} }},
\end{aligned}
\end{equation*}
where ${{\varepsilon _i}^{\left( {\bar F} \right)}}$ is the privacy budget of the $i$-th element of the noisy feature vector. Thus, we can get 
\begin{equation*}
    {M_F}\ln \left( {{\varepsilon _0}} \right) - {\varepsilon _0} \ge \ln \left( {\hat p} \right) + \sum\limits_{i = 1}^{{M_F}} {\ln \left( {{\Delta _i}} \right)}  + {M_F}\ln 2 .
\end{equation*}
From that, the total privacy budget $\varepsilon _0$ of the noisy feature vector has the ability to represent the accuracy of face recognition, \ie, equal total privacy budget $\varepsilon _0$ can be considered as equal face recognition accuracy. Therefore, we will no longer separately show the comparison of the face recognition accuracy of the five methods on the two data in this paper. 

\subsubsection{Evaluated Methods}
In experiments, we compare the proposed method with the existing advanced methods. And the methods are depicted as follows. 

\textbf{DCT-DP.} \cite{WOS:000897093900028} proposed a privacy-preserving face recognition method based on frequency-domain Discrete Cosine Transform (DCT) differential privacy, \ie, DCT-DP.

\textbf{Pixel-DP.} This method adds noises to all pixels of the face images. According to the geometric superposition, the Laplacian noises with the same scale parameter ${\bar b_P}$ are added to the selected geometric number of pixels of the face image, and the sum of the privacy budgets of the facial feature vector elements of the noisy image is the given total privacy budget.

\textbf{RDP.} This method is to add Laplacian noises with the same scale parameter ${\bar b_W}$, as shown in Definition \ref{def:wdp}. 

\textbf{RDP-NA.} This method is the optimized RDP with NA proposed in Sec. \ref{sec:na}). And its Laplacian noise with the optimized scale parameter ${{b_k}}$ is calculated by Eq. (\ref{eq:4.5}).

\textbf{RDP-LMGD.} The method is the optimized RDP through the LMGD proposed in Sec. \ref{sec:crc}). It calculates the optimized Laplacian noise scale parameter ${{b}_k}$ by LMGD, with the step length or learning rate set to $\eta \in [0.01, 0.1]$.

\subsection{Evaluation on Privacy Budget and Utility}
\subsubsection{Variance}

\begin{figure}[htbp]
	\centering
	\begin{minipage}{1\linewidth}	
		\subfigure[LFW Data]{
			\label{fig:variance_diff_epsilon_LFW}
			\includegraphics[width=0.49\linewidth]{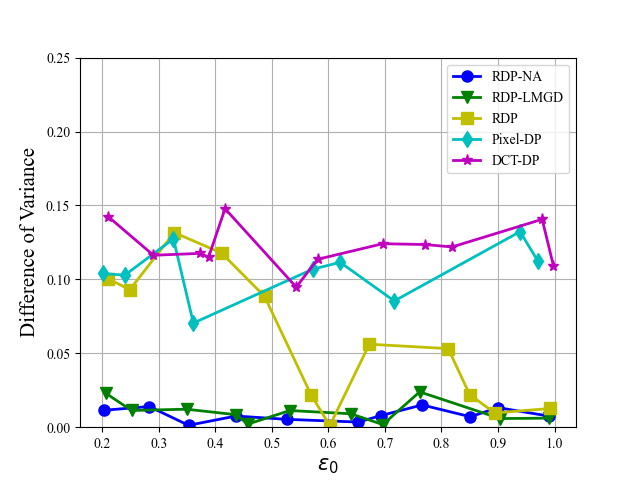}	
		}\noindent
		\subfigure[PubFig83 Data]{
			\label{fig:variance_diff_epsilon_PubFig83}
			\includegraphics[width=0.49\linewidth]{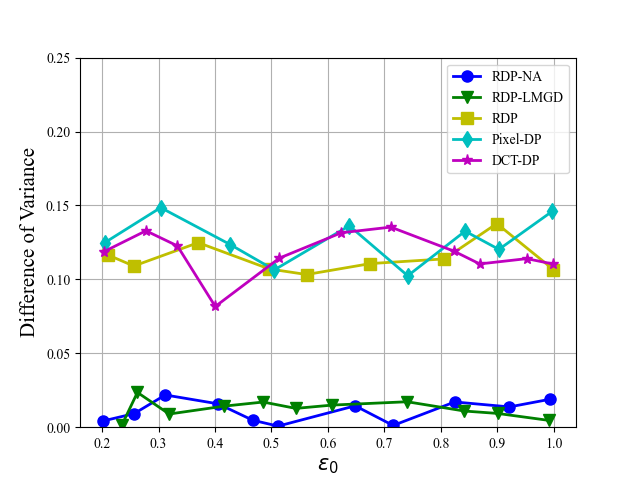}
		}
	\end{minipage}
	\caption{Normalized difference between real and theoretical variances, \ie, $\frac{{\left| {{\sigma _r}^2 - \sigma _t^2} \right|}}{{\sigma _t^2}}$.}
\end{figure}

	 


Now, let's look at the normalized difference or deviation between the real variance of numerical experiment $\sigma_r^2$ and the theoretical variance ${\sigma _t}^2$, which are shown in Figure \ref{fig:variance_diff_epsilon_LFW} and Figure \ref{fig:variance_diff_epsilon_PubFig83}. As the privacy budget $\varepsilon_0$ increased, for LFW data, the differences of the methods other than RDP (as shown in Definition \ref{def:wdp}) do not undergo drastic changes, but fluctuate within a narrow range; while the differences of the five methods consistently fluctuate within a small value range for PubFig83 data. Specifically, RDP-NA (RDP Optimized with Normalization Approximation, as seen in Sec.\ref{sec:na}) and RDP-LMGD fluctuate slightly near the minimum value of $0$ and do not exceed $0.05$ for LFW data and PubFig83 data. This demonstrates the effectiveness of our defined cost function for the privacy budget allocation optimization problem. In addition, among the normalized differences between the other three methods, RDP fluctuates greatly, but is never higher than $0.15$, while DCT-DP and Pixel-DP will never exceed $\left( {0.05, 0.15} \right)$ range. Obviously, our experiment is reasonable and valid

\begin{figure}[htbp]
	\centering
	\begin{minipage}{1\linewidth}	
		\subfigure[LFW Data]{
			\label{fig:variance_epsilon_LFW}
			\includegraphics[width=0.49\linewidth]{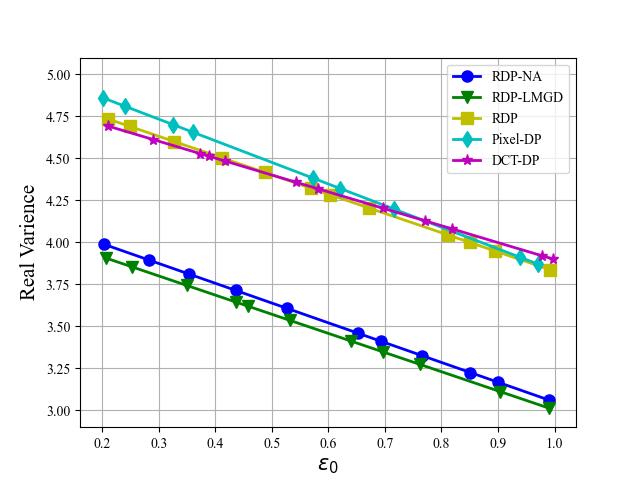}	
		}\noindent
		\subfigure[PubFig83 Data]{
			\label{fig:variance_epsilon_PubFig83}
			\includegraphics[width=0.49\linewidth]{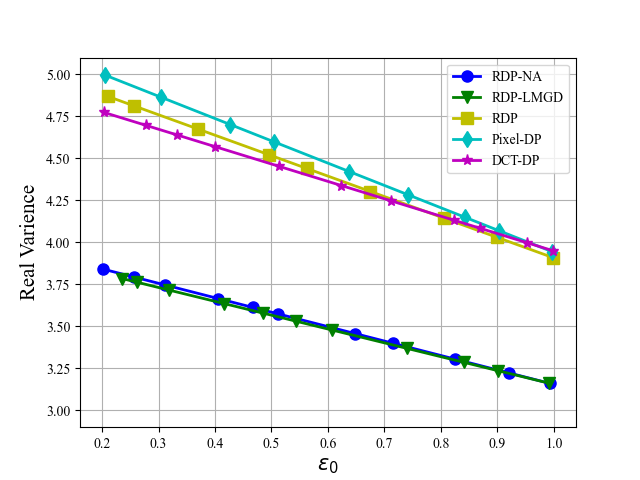}
		}
	\end{minipage}
	\caption{Logarithm of real variance $\log \left( {{\sigma _r}^2} \right)$  vs.  $\varepsilon_0$.}
\end{figure}


 

Finally, let's look at the real variance vs. the privacy budget $\varepsilon_0$. Figure \ref{fig:variance_epsilon_LFW} and Figure \ref{fig:variance_epsilon_PubFig83} show the fitted curves of the logarithm $\log \left({{\sigma _r}^2} \right)$ of the real variance $ {\sigma _r}^2$ of the five methods for the above two datasets under the given privacy budget. When the privacy budget $\varepsilon_0=0.2$,  the $\log \left({{\sigma _r}^2} \right)$ of RDP-NA and RDP-LMGD on both data is less than $4.0$. Meanwhile, $\log \left({{\sigma _r}^2} \right)$of DCT-DP is the smallest of the other three methods, around $4.75$, yet it is significantly higher than that of RDP-NA and RDP-LMGD, which shows that the real variance $ {\sigma _r}^2$ of RDP-NA and RDP-LMGD on the two data is significantly smaller than that of other methods when $\varepsilon_0=0.2$. Furthermore, as the privacy budget $\varepsilon_0$ increases, although the variance of all methods decreases significantly, no significant changes are found in this case, and $\log \left({{\sigma _r}^2} \right)$of RDP-LMGD is always slightly less than RDP-NA for LFW data, meaning that the real variance of RDP-LMGD sometimes is slightly less than RDP-NA. These findings suggest that RDP-NA and RDP-LMGD have less visual damage to the original face images than other methods.


\begin{table*}[htbp]
\centering
\caption{Privacy budget $\varepsilon_0$ vs. SSIM}
  \label{tab:ssim}
    \begin{tabular}{ccccccc}
\toprule
Datasets& Methods& ${\varepsilon _0} = 0.2$& ${\varepsilon _0} = 0.4$& ${\varepsilon _0} = 0.6$& ${\varepsilon _0} = 0.8$& ${\varepsilon _0} = 1.0$\\
\midrule
\multirow{5}{*}{LFW} & DCT-DP   &       $0.9767$&       $0.9908$&       $0.9937$&       $0.9956$&       $0.9965$\\ 
                          & Pixel-DP &       $0.9826$&       $0.9918$&       $0.9953$&       $0.9962$&       $0.9972$\\ 
                          & RDP&       $0.9844$&       $0.9926$&       $0.9953$&       $0.9963$&       $0.9973$\\ 
                          & RDP-NA&       $0.9954$&       $0.9980$&       $0.9988$&       $0.9992$&       $\textbf{0.9994}$\\ 
                          & RDP-LMGD&       $\textbf{0.9955}$&       $\textbf{0.9983}$&       $\textbf{0.9991}$&       $\textbf{0.9993}$&       $\textbf{0.9994}$\\ 
\midrule                
\multirow{5}{*}{PubFig83} & DCT-DP   &       $0.9728$&       $0.9882$&       $0.9931$&       $0.9947$&       $0.9958$\\ 
                          & Pixel-DP &       $0.9836$&       $0.9923$&       $0.9951$&       $0.9964$&       $0.9971$\\ 
                          & RDP&       $0.9815$&       $0.9905$&       $0.9940$&       $0.9960$&       $0.9969$\\ 
                          & RDP-NA&       $\textbf{0.9968}$&       $0.9985$&       $0.9989$&       $0.9992$&       $\textbf{0.9994}$\\ 
                          & RDP-LMGD&       $0.9967$&       $\textbf{0.9986}$&       $\textbf{0.9991}$&       $\textbf{0.9993}$&       $\textbf{0.9994}$\\
\bottomrule
\end{tabular}
\end{table*}

\subsubsection{SSIM}

In this paper, we conduct simulation experiments of SSIM on two real datasets for five methods within a given privacy budget range, as shown in Table \ref{tab:ssim}. It can be seen that with the increase of privacy budget $\varepsilon_0$, RDP-NA and RDP-LMGD always have higher SSIM than other methods. This indicates that the scale parameters of Laplacian noise optimized by NA or LMGD can make the noisy image more similar to the original face image. In particular, the SSIM of RDP-LMGD is always slightly higher than that of RDP-NA in most cases, except that SSIM of RDP-NA in LFW data is slightly higher than that of RDP-LMGD when ${\varepsilon _0} = 0.2$, and SSIM of RDP-NA and RDP-LMGD are equal when ${\varepsilon _0} = 1.0$ on both data. 

\subsubsection{PSNR}
Finally, let's look at the visualization quality of the sanitized face images after the sanitizing noise is added. 

\begin{figure}[htbp]
	\centering
	\begin{minipage}{1\linewidth}	
		\subfigure[LFW Dataset]{
			\label{fig:psnr_LFW}
			\includegraphics[width=0.49\linewidth]{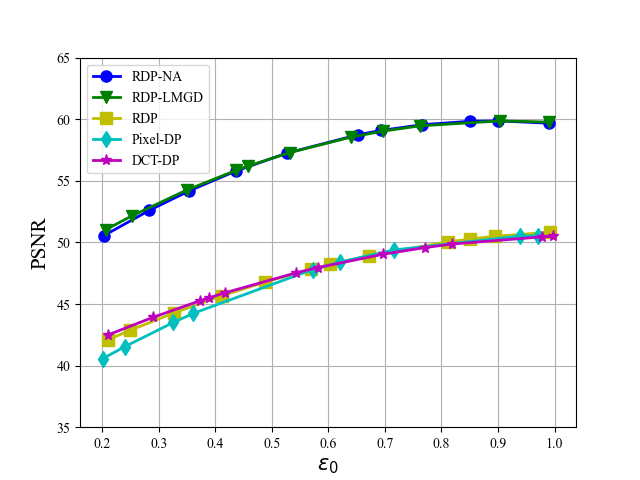}	
		}\noindent
		\subfigure[PubFig83 Dataset]{
			\label{fig:psnr_PubFig83}
			\includegraphics[width=0.49\linewidth]{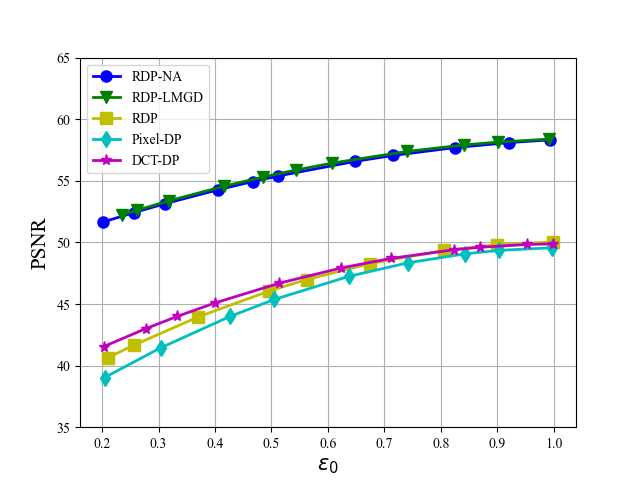}
		}
	\end{minipage}
	\caption{Privacy Budget $\varepsilon_0$ vs. PSNR.}
	\label{fig:psnr}
\end{figure}

In this paper, we compared the PSNR of the noise sanitized face images produced by the tested methods within a given range of total privacy budgets from $\varepsilon_0 = 0.2$ to $\varepsilon_0 = 1.0$ with the geometric superposition of parameter $p=0.02$. We obtained face images from two datasets within the given privacy budgets. The fitting curves of PSNR vs. face images are shown in Figure \ref{fig:psnr}, from which we can observe that as the privacy budget increases, the PSNR becomes higher. As shown in Figure \ref{fig:psnr_LFW}, the PSNR of RDP-NA and RDP-LMGD for LFW data higher than all compared methods. In particular, they are very close. Also, as the privacy budget increases, the advantages of RDP-NA and RDP-LMGD become more pronounced. Similarly, Figure \ref{fig:psnr_PubFig83} shows the result For PubFig83 data, from which it is clear that the PSNRs of both RDP-NA and RDP-LMGD are much higher than those of compared  methods. For both data at high level of noise of $\varepsilon_0 = 0.2$, the PSNRs of both RDP-NA and RDP-LMGD significantly exceeded $50$dB, while the best PSNR of the compared methods is only around $42$dB, which is $\sim$$10$dB less than that of RDP-NA. In sum, both RDP-NA and RDP-LMGD achieve better visualization quality of the sanitized face images. 

\begin{figure}[htbp]
	\centering
	
		\subfigure[Original face image]{
			\label{fig:face_original_LFW}
			\includegraphics[width=0.25\linewidth]{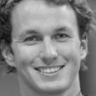}	
		}\noindent
        \subfigure[Dct-DP]{
			\label{fig:face_dctdp_LFW}
			\includegraphics[width=0.25\linewidth]{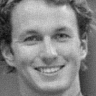}
			
		}\noindent
    \subfigure[Pixel-DP]{
			\label{fig:face_pixeldp_LFW}
			\includegraphics[width=0.25\linewidth]{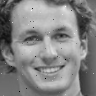}	
		}\noindent\\
  \subfigure[RDP]{
			\label{fig:face_RDP_LFW}
			\includegraphics[width=0.25\linewidth]{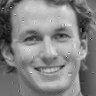}
		}\noindent
		\subfigure[RDP-NA]{
			\label{fig:face_RDPna_LFW}
			\includegraphics[width=0.25\linewidth]{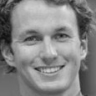}
			}
		\noindent
  \subfigure[RDP-LMGD]{
			\label{fig:face_RDPLMGD_LFW}
			\includegraphics[width=0.25\linewidth]{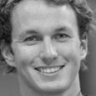}
			}
	
	\caption{Visualization of LFW Data.}
	\label{fig:face_visiual_LFW}
\end{figure}

\begin{figure}[htbp]
	\centering
	
		\subfigure[Original face image]{
			\label{fig:face_original_PubFig83}
			\includegraphics[width=0.25\linewidth]{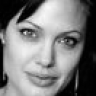}	
		}\noindent
        \subfigure[Dct-DP]{
			\label{fig:face_dctdp_PubFig83}
			\includegraphics[width=0.25\linewidth]{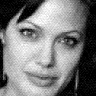}
			
		}\noindent
    \subfigure[Pixel-DP]{
			\label{fig:face_pixeldp_PubFig83}
			\includegraphics[width=0.25\linewidth]{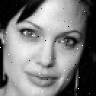}	
		}\noindent\\
  \subfigure[RDP]{
			\label{fig:face_RDP_PubFig83}
			\includegraphics[width=0.25\linewidth]{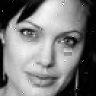}
		}\noindent
		\subfigure[RDP-NA]{
			\label{fig:face_RDPna_PubFig83}
			\includegraphics[width=0.25\linewidth]{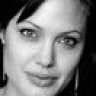}
			}
		\noindent
  \subfigure[RDP-LMGD]{
			\label{fig:face_RDPLMGD_PubFig83}
			\includegraphics[width=0.25\linewidth]{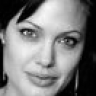}
			}
	
	\caption{Visualization of PubFig83 Data.}
	\label{fig:face_visiual_PubFig83}
\end{figure}

Now, let's look at the visualization effect. As shown in Figure \ref{fig:face_visiual_LFW} and Figure \ref{fig:face_visiual_PubFig83}, we show the sanitized face images of five methods obtained  with a privacy budget $\varepsilon_0 = 0.2$ and the geometric superposition parameter of $p=0.02$ for LFW data and FubFig83 data. It can be shown that the Pixel-DP obtains more local mosaics on images than other wavelet transform based methods, which means that adding noise to selective wavelet coefficients of large significance on the privacy budget is better than adding noise directly to image pixels. Also, compared to the other four methods, the DCT-DP method contains deviation cross the whole face image, which is due to the global effect of the cosine basis used. On the contrary, no significant distortion nor visable mosaic effect can be seen in the sanitized face images of both RDP-NA and RDP-LMGD,  which means better data utility of visualization quality of the sanitized face images has been achieved.

\subsection{Evaluation of Bypass Attacks Against}

We use the open-source facial recognition library face-recognition to calculate the Euclidean distance of the encoding vectors of the noisy facial image and the original face image. If the Euclidean distance is less than the set threshold $0.975$, it is considered that the noisy facial image belongs to the certain individual to whom the original facial image corresponds. Otherwise, we consider it not the same face as the original facial image. 

Table~\ref{tab:error rate} shows the false negative rate of the face recognition corresponding to the five methods under the same utility settings on LFW data.

\begin{table}[htbp]
\centering
\caption{False Negative Rate under the Same Utilty Setting}
\label{tab:error rate}
\begin{tabular}{cccccc}
\toprule
Dataset&Method&False Negative Rate\\
\midrule                
\multirow{5}{*}{LFW}&DCT-DP&$37.75\%$\\ 
&Pixel-DP&$39.44\%$\\ 
&RDP&$40.05\%$\\
&RDP-NA&$77.18\%$&\\ 
&RDP-LMGD&$\textbf{81.38\%}$\\ 
\bottomrule
\end{tabular}
\end{table}

\section{Conclusion}
In this paper, a novel and efficient RDP is proposed to protect the facial features privacy for face image data, which can be formulated as the constraint optimization problem of maximizing the data utility or visualization quality of the noise sanitized face images, for a given privacy budget $\varepsilon_0$. First, the RDP calculates the influence or weights of the noisy feature coefficients in the multiscale transform subspaces with respect to the total privacy budget $\varepsilon_0$; then, the RDP adds Laplacian noises to the weight-ranked feature coefficients by geometric superposition, followed by the inverse transform to obtain the noise sanitized face image. It is rigorously proved that the noisy facial feature vector obtained with RDP satisfies $\varepsilon_0$-differential privacy. After that, the constraint optimization problem is formulated by the LM method by making the first-order Taylor expansion approximation on the influence or weight of the feature coefficients on the facial features. Furthermore, to solve the nonlinear LM, two methods are proposed, one for the real-time online applications and the other for the accurate offline applications: 1) the analytical NA method  for the real-time online applications: by assuming that all Laplacian noise scale parameters to be identical to their average value;  and 2) the LMGD method for the accurate offline applications: by computing the optimal scale parameters of Laplacian noise through iterative updating. Experiments on two real-world datasets are performed, which show that the RDP does achieve better data utility for a given facial feature privacy budget $\varepsilon_0$, compared to other state-of-the-art DP methods. 

\newpage
\bibliographystyle{IEEEtran}
\bibliography{IEEEexample}
%



\end{document}